\documentclass[letterpaper, 10 pt, conference]{ieeeconf}  

\IEEEoverridecommandlockouts                              

\usepackage[linesnumbered,ruled,vlined]{algorithm2e}
\usepackage[T1]{fontenc}
\usepackage{algorithmic}
\usepackage{graphics} 
\graphicspath{{figures/}}
\usepackage{epsfig} 
\usepackage{times} 
\usepackage{siunitx}
\usepackage{amsmath} 
\usepackage{amssymb}  
\usepackage[colorlinks = true,
            linkcolor = black,
            urlcolor  = black,
            citecolor = black,
            anchorcolor = black]{hyperref}
\usepackage[capitalise]{cleveref}
\Crefname{algorithm}{Alg.}{Algs.}
\Crefname{section}{Sec.}{Secs.}
\Crefname{equation}{Eq.}{Eqs.}
\usepackage{tikz}
\usetikzlibrary{positioning, shapes.geometric, arrows}
\usepackage{booktabs}
\usepackage{tabularx}
\usepackage{multirow}
\usepackage{wrapfig}

\newcolumntype{Y}{>{\centering\arraybackslash}X} 

\usepackage[
    sorting=none,
    giveninits=true,
    maxbibnames=9,
    maxcitenames=2,backend=biber
    ]{biblatex}
\usepackage{etoolbox}
\makeatletter
\patchcmd{\@makecaption}
  {\scshape}
  {}
  {}
  {}
\makeatletter
\patchcmd{\@makecaption}
  {\\}
  {.\ }
  {}
  {}
\makeatother

\renewcommand{\baselinestretch}{0.99}

\title{\LARGE \bf
UPTor: Unified 3D Human Pose Dynamics and Trajectory Prediction for Human-Robot Interaction
}

\author{
Nisarga Nilavadi$^{1,2}$, Andrey Rudenko$^{1}$, Timm Linder$^{1}$%
\thanks{$^{1}$Bosch Corporate Research, Robert Bosch GmbH, Stuttgart, Germany {\tt\small andrey.rudenko@bosch.com}}
\thanks{$^{2}$University of Technology Nuremberg (UTN), Germany}%
\thanks{$^{3}$This work has received funding from the European Union’s Horizon 2020 research and innovation programme under grant agreement No 101017274 (DARKO). \url{https://darko-project.eu/}}
}

\begin{document}

\maketitle
\thispagestyle{empty}
\pagestyle{empty}

\begin{abstract}
We introduce a unified approach to forecast the dynamics of human keypoints along with the motion trajectory based on a short sequence of input poses. While many studies address either full-body pose prediction or motion trajectory prediction, only a few attempt to merge them. We propose a motion transformation technique to simultaneously predict full-body pose and trajectory key-points in a global coordinate frame. We utilize an off-the-shelf 3D human pose estimation module, a graph attention network to encode the skeleton structure, and a compact, non-autoregressive transformer suitable for real-time motion prediction for human-robot interaction and human-aware navigation. We introduce a human navigation dataset ``DARKO'' with specific focus on navigational activities that are relevant for human-aware mobile robot navigation. We perform extensive evaluation on Human3.6M, CMU-Mocap, and our DARKO dataset. In comparison to prior work, we show that our approach is compact, real-time, and accurate in predicting human navigation motion across all datasets. Result animations, our dataset, and code will be available at \url{https://nisarganc.github.io/UPTor-page/}
\end{abstract}


\section{Introduction}
Human motion forecasting and activity recognition is a critical component for social robots and autonomous systems that need to operate and interact with people in domestic and industrial environments \cite{rudenko2020human}. Accurate trajectory prediction in crowded environments has strong impact on the effectiveness and safety of a mobile robot, allowing smooth and unobtrusive navigation \cite{heuer2023proactive}. Further refining trajectory predictions with full-body poses provides complete information about human behaviour with many promising applications in human-robot interaction \cite{mahdavian2023stpotr, butepage2018anticipating}, automated driving \cite{mangalam2020disentangling}, surveillance \cite{li2023future} and healthcare \cite{kidzinski2020deep}.  

\textbf{Trajectory Prediction} \cite{alahi2016social,Gupta2018SocialGAN,amirian2019social,zhao2019multi,salzmann2020trajectronpp,wang2023anyposeanytime3dhuman} focuses on the coarse-level root joint trajectories of vehicles, pedestrians or other dynamic agents, dealing with broader movement patterns in a global coordinate space \cite{rudenko2020human}. \textbf{Full-body Pose Prediction}, also referred to as \textbf{Human Motion Prediction} \cite{ma2022progressivelygeneratingbetterinitial, guo2022mlpsimplebaselinehuman, martinez2021pose, wang2021multiperson, giuliari2021transformer}, on the other hand, deals with the fine-grained prediction of 3D skeleton joints coordinates, relative to the fixed root joint (e.g. hip or torso) resulting in pose prediction in a local coordinate space \cite{mahdavian2023stpotr} without considering the global translation. Historically speaking, the research in trajectory prediction and full-body pose prediction, with a few notable exceptions \cite{cao2020long, mahdavian2023stpotr, nikdel2023dmmgan}, progressed independently and targeted distinct application scenarios. However, solving these tasks in a unified manner is advantageous for effective human-robot interaction. Incorporating the full-body pose features can refine dynamics modeling in trajectory prediction \cite{unhelkar2015human,rudenko2020human}. Similarly, full-body pose prediction in global coordinate space aids robot response in approach and handover applications, allowing to better plan the avoidance maneuvers in close proximity \cite{stefanini2024ral}.

Prior art addressed the problem of pose and trajectory prediction as two separate tasks \cite{yuan2020dlow, martinez2021pose} or solved it in a decoupled manner with separate modules for each task \cite{parsaeifard2021learning, mahdavian2023stpotr}. Furthermore, only a handful of works \cite{mahdavian2023stpotr} specifically focused on navigation activities, which are of critical interest for predictive planning of a mobile robot, over more diverse actions without distinct locomotion. This research gap is also reflected in the prior art datasets of full-body motion, which are often dominated by static activities such as bending, reaching, handing over, standing up, etc.

\begin{figure}[t]
\centering
\includegraphics[width=\linewidth]{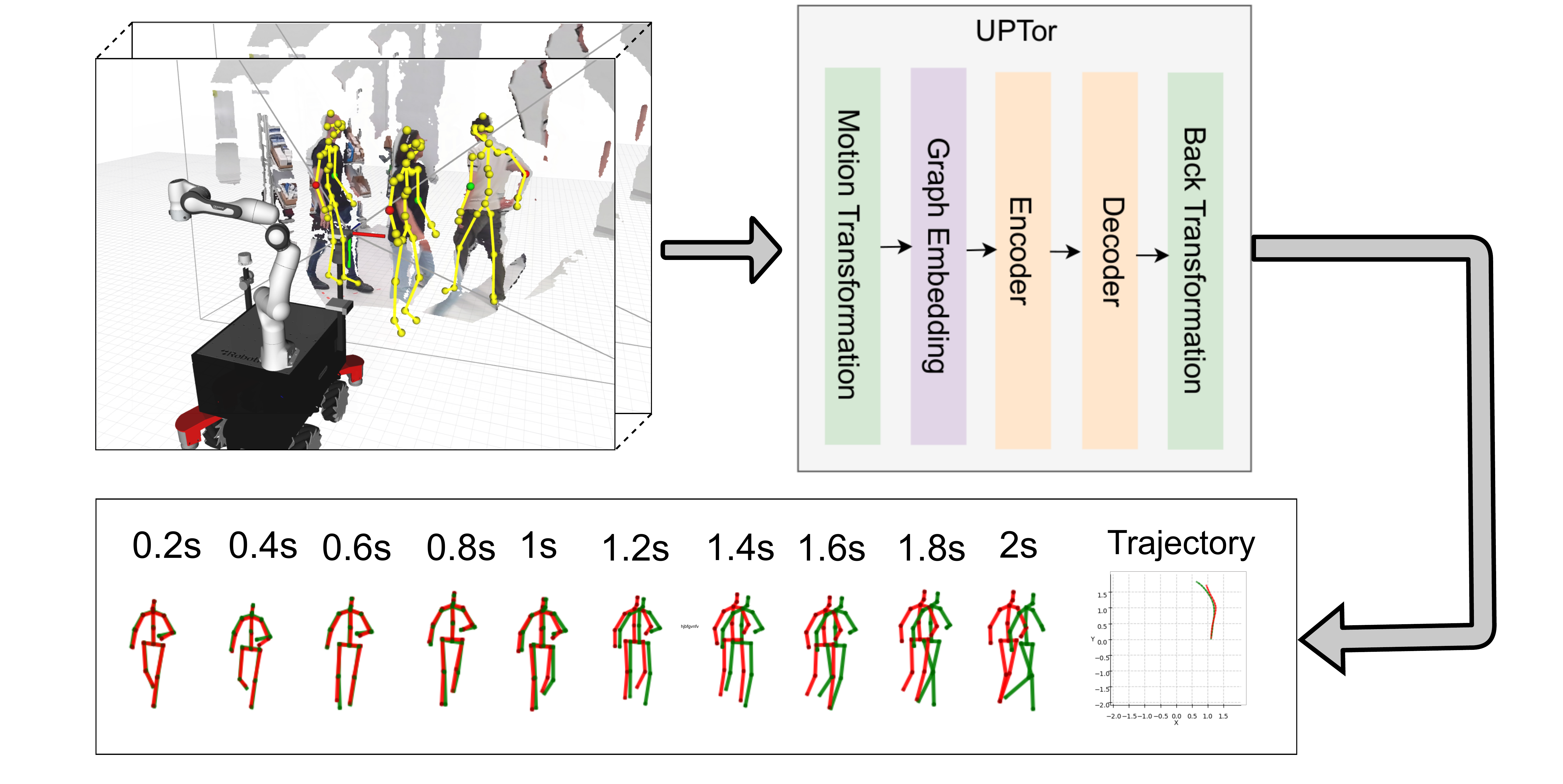}
\vspace*{-5mm}
\caption{Human key points estimated from the robot’s perception stack and human motion prediction in global 3D coordinates from our UPTor model.} 
\label{fig:cover}
\vspace*{-4mm}
\end{figure}

In this work we develop an approach for Unified human Pose and Trajectory prediction (``UPTor'') for real-time and safe human-aware motion planning of our intralogistics robot$^{3}$, see Fig.~\ref{fig:cover}. We encode the skeleton features with a graph attention network and utilize a non-autoregressive Transformer model \cite{vaswani2017attention, martinez2021pose} to process input pose sequences in 3D global coordinates. To support this form of input and output, we propose a motion transformation technique and train the transformer on sequences within the transformed coordinate space. We evaluate our method on the H3.6M \cite{h36m_pami} and CMU-Mocap \cite{cmumocap} datasets, and contribute a novel DARKO dataset of 17 subjects performing 508 sequences of diverse navigation-related activities. Through qualitative and quantitative experiments, we demonstrate that our approach is compact, fast, and accurate in full-body pose and trajectory prediction of navigating humans for robotic applications.

In summary, we make the following contributions:
\begin{itemize}
    \item Human motion prediction with global translation in global 3D coordinates using our motion transformation technique, which can be applied to any local pose prediction method to achieve the rare and unexplored unified task of pose and trajectory prediction;
    \item A compact, accurate, and real-time human navigation prediction approach that is well-suited for robotics applications;
    \item A human navigation dataset with diverse actors collected using the 3D perception stack of a mobile robot.
\end{itemize}

\section{Related Work}
\label{sec:related_work}

Many human motion prediction methods in the computer vision community focus on accurate root-relative pose predictions and excel at generating poses in a stochastic manner within local space \cite{yuan2020dlow, barquero2023belfusionlatentdiffusionbehaviordriven, tian2023transfusionpracticaleffectivetransformerbased, wang2021simple}. Similarly, many trajectory prediction studies from autonomous systems community focus on predicting plausible trajectories in unstructured environment \cite{Gupta2018SocialGAN,amirian2019social}. Predicting human poses and trajectories together to forecast complete human motion in global space is a relatively new but actively developing research direction \cite{mangalam2020disentangling,parsaeifard2021learning,wang2021simple,mahdavian2023stpotr,nikdel2023dmmgan,adeli2020socially,cao2020long,adeli2021tripod}. Traditionally, RNNs were used to generate sequential predictions \cite{fragkiadaki2015recurrent, martinez2017human}. Yet, their inherent nature of processing data sequentially can be a bottleneck, especially in real-time applications. Aiming for a fast and accurate solution for applications on a mobile robot, in this work we explore a non-autoregressive variant of Transformer, which performs simultaneous decoding of the entire output sequence in a single step, as in the Pose Transformer (POTR)~\cite{martinez2021pose}. POTR is specifically designed for root-relative pose prediction and thus removes global translation from human motion sequences. STPOTR \cite{mahdavian2023stpotr} extends POTR with a second transformer for trajectory prediction and introduces shared attention layers between the pose and trajectory transformers. Since STPOTR predicts poses and trajectory in a decoupled fashion, 3D human poses are pre-processed to separate the global movement from poses by subtracting pose dynamics joints from the root joint. Predicted pose and trajectory key-points are later concatenated to obtain the global pose over time. 
DeRPoF \cite{parsaeifard2021learning} also adopts a decoupled approach with an LSTM-based encoder-decoder for trajectory forecasting and a Variational Autoencoder (VAE) encoder-decoder to predict root-relative pose dynamics.
DMMGAN \cite{nikdel2023dmmgan} is an accurate but computationally-expensive model designed for predicting diverse human motions, including both trajectories and poses, using a Transformer-attention based network.
Diversifying Latent Flows (Dlow) \cite{yuan2020dlow} aims to generate multiple hypotheses of human poses utilizing a pre-trained deep generative model.

Our approach directly utilizes the 3D joint positions from an off-the-shelf 3D human pose estimation module \cite{sarandi2020metrabs}. We propose a novel input transformation technique to couple pose and trajectory prediction, thus simplifying the model and improving the runtime. We also leverage graph attention networks to generate graph embedding that captures spatial skeleton structure, thus informing the model of the skeletal hierarchy and relative dependencies between individual joints. Finally, we predict human pose dynamics and trajectory with a non-autoregressive transformer, optimizing computational efficiency for real-time robotic predictions.

Our motion transformation technique deals with the entire length of the learning sequence (i.e. observed motion concatenated with the ground truth prediction) in global 3D coordinates, temporally aligned in position and orientation around the positive x-axis of global coordinate space. This differs from aligning the root joint of each individual pose, as it is common in the pose prediction methods \cite{xu2023eqmotionequivariantmultiagentmotion}, the beginning of the observed motion as we found in the code of the related approach \cite{tevet2022humanmotiondiffusionmodel}, or with respect to the pose of the target object as in \cite{xu2023interdiffgenerating3dhumanobject}. Training data, transformed according to our method, can naturally achieve the rare and useful task of joint pose and trajectory prediction for mobile robots.

\section{Methodology}\label{sec:method}

\subsection{Problem Formulation}\label{sec:problem}
Let \( \mathcal{P}(t) \in \mathbb{R}^{3N} \) denote the 3D human pose at time $t$ comprising $N$ joints: $\mathcal{P}(t) = \{j_1(t), \dots, j_N(t)\}
$ where each \(j_i(t) \in \mathbb{R}^3\) represents the $(x, y, z)$ coordinates of the \(i^{\text{th}}\) joint in the global coordinate frame at time \(t\). We define an input sequence as a set of poses from time $1$ to time $T_1$: $\mathcal{S}_{\text{in}} = \{ \mathcal{P}(1), \dots, \mathcal{P}(T_1) \in \mathbb{R}^{T_1 \times 3N}$. The objective of the model is to predict sequence of poses from $T_1+1$ to $T_1+T_2$ in global coordinate frame: $\mathcal{S}_{\text{out}} = \{ \mathcal{P}(T_1+1), \dots, \mathcal{P}(T_1+T_2) \} \in \mathbb{R}^{T_2 \times 3N}$. The complete motion sequence $\mathcal{S}$ is given by $\mathcal{S} = \mathcal{S}_{\text{in}} \cup \mathcal{S}_{\text{out}}$.

\begin{figure*}[!t]
\centering
\vspace*{-4mm}
\includegraphics[angle=270, width=0.97\linewidth]{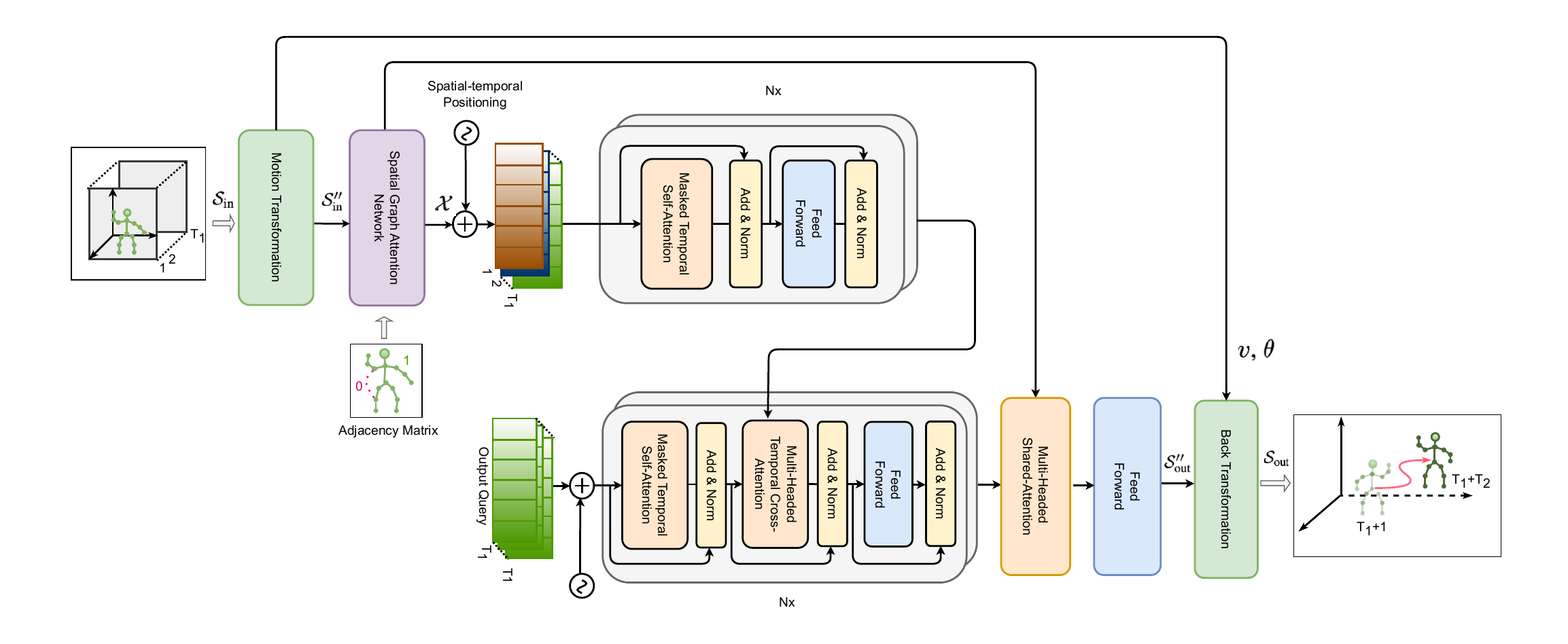}
\vspace*{-5mm}
\caption{UPTor: Unified 3D Human Pose Dynamics and Trajectory Prediction Transformer}
\label{fig:network}
\end{figure*}

\subsection{Model Architecture}
\begin{figure}[t]
  \begin{center}
    \includegraphics[width=\linewidth,trim={0 12mm 0 0},clip]{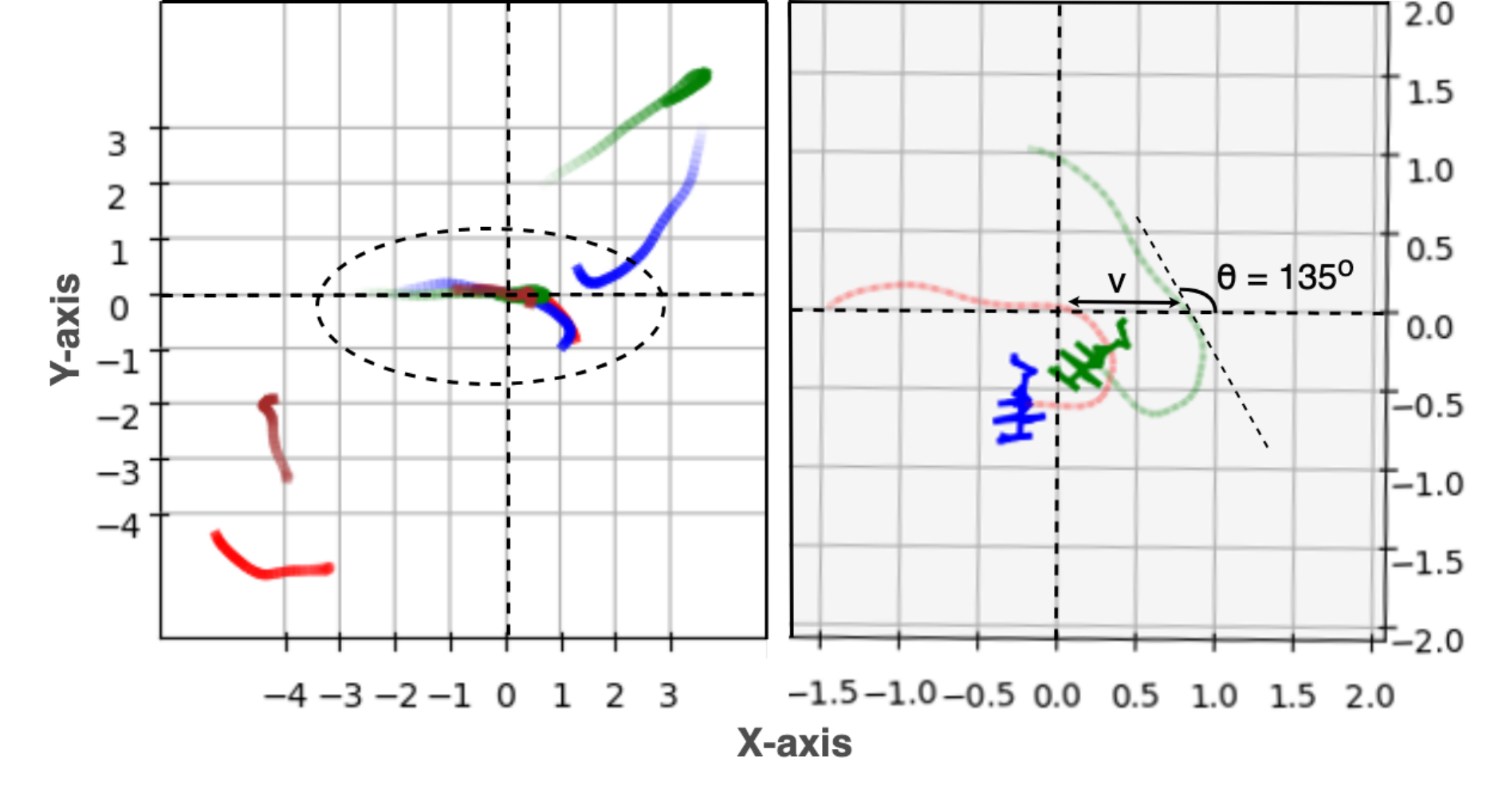}
  \end{center}
  \vspace*{-3mm}
\caption{\textbf{Left:} A top-down view of four color-coded motion sequences, with their corresponding transformed sequence encircled. Faded colors mark the motion start. \textbf{Right:} Transformation of a single sequence. Original motion is represented by a green trajectory and skeleton. Transformation parameters include the angle $\theta$ between the motion direction and the positive x-axis, and the translation vector $v$ at \(T_1\).}
\label{fig:transfromation}
\vspace*{-1mm}
\end{figure}

\textbf{Motion Transformation:} We first address the challenge of training motion sequences from global coordinate frame with varying initial positions and motion orientations. To that end, we propose a systematic method to normalize the motion sequences and achieve global and orientation invariance, as summarized in Fig.~\ref{fig:transfromation}.

\textit{Global invariance:} To ensure that our predictions commence from consistent coordinates, we translate the entire motion sequence using the vector \( {v} \), derived as the negative counterpart of the root joint position at the last pose of the input sequence \( j_{\text{root}}(T_1) \) i.e., ${v} = - j_{\text{root}}(T_1)$. We add this translation vector to every pose in the sequence, as:
\begin{equation}
\mathcal{P}'(t) = \mathcal{P}(t) + {v} \quad \forall t \in [1, T_1 + T_2]
\end{equation}

\textit{Orientation invariance:} To align various motion directions along with the positive x-axis, the rotation angle, \( \theta \) is computed in radians between the input motion direction and the positive x-axis. This angle is calculated as the arc tangent of the ratio of the differences in the \( y \) and \( x \) coordinates of the root joint's position at the last input pose \( T_1 \) and the pose at \( (T_1- \delta ) \), with \( \delta \) being a predetermined interval to measure motion direction at \( T_1 \).
\vspace*{-2mm}
\begin{equation}
\theta = \arctan2(\Delta y, \Delta x) 
\left\{ 
\begin{aligned}
\Delta x &= j_{\text{root}}(T_1)_x - j_{\text{root}}(T_1-\delta)_x \\
\Delta y &= j_{\text{root}}(T_1)_y - j_{\text{root}}(T_1-\delta)_y 
\end{aligned}
\right.
\end{equation} 

Finally, to rotate the entire sequence, we take the dot product of the rotation matrix around the z-axis, \( R_z \) with all translated poses $\mathcal{P}'(t)$, achieving the transformed sequence:
\vspace*{-1.5mm}
\begin{align*}
\mathcal{S}'' = & \{\mathcal{P}''(t)\}, \text{where } \mathcal{P}''(t) = \\
& \begin{bmatrix}
   \cos(-\theta) & -\sin(-\theta) & 0 \\
   \sin(-\theta) & \cos(-\theta) & 0 \\
   0 & 0 & 1
   \end{bmatrix}
\cdot \mathcal{P}'(t) \quad \forall t \in [1, T_1 + T_2]
\end{align*}

Motion Transformation allows our model to couple the pose and trajectory prediction into a single problem, thereby avoiding the separation of root joint movement from other pose dynamics joints as in \cite{nikdel2023dmmgan,parsaeifard2021learning,mangalam2020disentangling}, which leads to unnatural decoupling of the articulated poses from global motion as noted by \cite{adeli2020socially}.

\textbf{Spatial Graph Embedding:}\label{sec:sge} We utilize a Graph Attention Network (GAT) \cite{velivckovic2017graph} to generate graph embedding for each pose $\mathcal{P}''(t)$ in the input sequence. 
To that end, we represent the pose as a graph, where each joint corresponds to a node and bones to the edges. Input to the GAT module is the reshaped sequence, $\mathcal{S}_{\text{in}}'' \in \mathbb{R}^{T_1 \times N \times 3}$, where each node \(j_i(t) \in \mathbb{R}^3\) has 3 spatial features. Edges, \( E \) are determined based on the kinematic chain of the skeleton model, depending on the dataset. From this kinematic chain, we derive an adjacency matrix \( A \) to represent connections between joints such that \( A_{ij} = 1 \) if there is a connection between joint \( j_i(t) \) and \( j_j(t) \), and 0 otherwise.
The GAT computes attention scores between joint pairs, to capture the spatial attention between joints of the same frame.
The joint features are updated by aggregating features from neighboring joints, resulting in a joint embedding.

The GAT layer produces a new set of node features \( \mathcal{X} \in \mathbb{R}^{T_1 \times N \times J_{\text{dim}}} \) as its output, yielding the joint embedding. Subsequently, joint embedding from the same pose are flattened to create the pose embedding, which has a dimension of \( \mathcal{X} \in \mathbb{R}^{T_1 \times (N \times J_{\text{dim}}) } \). Thus, the dimensionality \(D\) of the transformer model is given by \( N \times J_{\text{dim}} \). The GAT here is employed to facilitate intra-frame attention mechanisms among joints, effectively capturing the spatial relationships. The output from the GAT, which represents spatial embedding, is subsequently fed into a transformer module. The transformer is designed to learn temporal relations across frames, ensuring a comprehensive understanding of both spatial and temporal dynamics of human motion. 

\textbf{Spatial-Temporal Positioning:} We incorporate dual layers of positional encoding to capture human dynamics in detail, building on the formulation from \cite{vaswani2017attention}. First we generate a sinusoidal spatial positional encoding to establish differentiation amongst various joints within each pose. For every joint, this method produces a positional encoding of dimension \( J_{\text{dim}} \), accounting for all \( N \) joints. Subsequently, to differentiate between poses over time, we generate a temporal encoding. Temporal encodings have dimension \( J_{\text{dim}} \times N \), accounting for all \( T_1 \) poses, elucidating the sequential dynamics from one frame to the next.

\textbf{Transformer Encoder Decoder:} The basic structure of the Transformer layers is adopted from \cite{vaswani2017attention} with a non-autoregressive decoder inspired by POTR \cite{martinez2021pose}. The transformer takes the spatio-temporally positioned input poses and processes them through a number of Nx encoder and decoder layers. Each layer of encoder and decoder, shown in Fig.~\ref{fig:network}, incorporates a Temporal Self-Attention component adopted from \cite{shaw2018selfattention} that emphasize the relative distances between tokens in a sequence. In this technique, for each pose, attention scores are weighted more heavily towards its immediate neighboring poses. This is particularly beneficial for human pose sequences, where not only the order of positions are crucial, but the relative transition between frames is also extremely important. In addition, to ensure that the human pose at current step is dependent only on prior poses and not on any future poses, we employ causal masking to Temporal Self-Attention components.  

The output from encoder block, which projects the input sequence into a latent space \( Z = [z_1, \dots, z_T] \), is passed to the decoder. Decoder queries are initialised with \( \mathcal{X}(T_1) \) which is the encoder's last input pose repeated over target length times. 
Afterwards, a Multi-Headed Shared-Attention Mechanism is employed, wherein the decoder output query attends to the output of the Graph Attention Network, \( \mathcal{X} \). Subsequently, these output embedding are propagated through linear layers and then transformed back to their original motion orientation and global coordinate space using \( \vec{v} \) and \( \theta \) resulting in the output sequence \(\mathcal{S}_{\text{out}}\).

Our model was trained using the L2-norm loss between predicted poses \( \hat{P}_{t} \) and the ground truth poses \( P_{t} \) across the prediction horizon:
\vspace*{-2mm}
\begin{align*}
L = \frac{1}{T_2 - T_1 - 1} \sum_{t = T_1 + 1}^{T_1 + T_2} \|\hat{P}_t - P_t\|_2 
\end{align*}


\subsection{Implementation Details}
The proposed architecture is implemented using the PyTorch deep learning framework. The model is trained with AdamW optimizer \cite{loshchilov2019decoupled} for a maximum epochs of 20 on Human3.6M dataset, 125 epochs on DARKO dataset and 50 epochs on CMU-Mocap dataset. Learning rate is set to \(10^{-5}\) and weight decay to \(10^{-5}\). For all the datasets, we consistently set \(J_{\text{dim}}\) to 32. Consequently, the model dimensions \(D\) are 544 for Human3.6M, 992 for CMU-Mocap, and 960 for DARKO dataset, varying with the number of joints in the dataset as detailed in Sec.~\ref{sec:experiment_setup}.

\section{Experiments}
\label{sec:experiments}

\subsection{Setup}
\label{sec:experiment_setup}

\textbf{Datasets:} We evaluate our approach on the common public datasets, Human3.6M \cite{h36m_pami} and CMU-Mocap \cite{cmumocap}, and on our new DARKO dataset. We classify the human motion sequences from these datasets into two main action categories: navigational motions, such as walking, running, or greeting, and static activities without distinct locomotion, such as discussing, smoking, or eating.

The \textbf{Human3.6M} dataset \cite{h36m_pami} consists of 17 distinct actions, each performed by 11 subjects, resulting in 3.6 million frames. The dataset features 32 human joints recorded at 50 Hz using a motion capture system. Similar to the prior art evaluation \cite{mahdavian2023stpotr, yuan2020dlow, nikdel2023dmmgan}, we reduced the frame rate to 10 Hz which is suitable for robotic applications, adopted a 17-joints skeleton model with 16 joints capturing the pose dynamics, and a root joint representing the trajectory. We conduct training on five subjects S1, S5, S6, S7, S8 (389,938 frames), and test on two subjects S9 and S11 (135,836 frames), each subject performing 15 specific actions: Directions, Discussion, Walking, WalkTogether, Sitting, Smoking, Waiting, Posing, Purchases, Greeting, Eating, Phoning, SittingDown, Photo, and WalkDog. The input spans 0.5 \si{\sec} (5 frames), and the output covers 2.0 \si{\sec} (20 frames).

The \textbf{CMU-Mocap} \cite{cmumocap} is an extensive human motion database with diverse activities, ranging from common actions like walking and running to unusual activities such as animal mimicry, dancing, or swimming. 
In our study, we focused on locomotion-related activities. We filtered out sequences labeled with keywords such as ``walk'', ``run'', ``jump'', and ``navigate'' and curated a subset that highlights various locomotion patterns, including running, walking, jogging, and navigating. This extracted data includes sequences from subjects 2 through 141, each performing different activities with multiple trials for each activity. The training set consists of 165,677 frames from all sampled subjects, while the test set contains 12,854 frames of unseen trials of activities from these subjects. We reduced the frame rate to standard 10 Hz. This dataset exhibits unpredictable, dramatic, and dynamic trajectories, so we conduct short-term pose and trajectory predictions. The input horizon is 0.5 \si{\sec} (5 frames), and the output horizon 1 \si{\sec} (10 frames).

\begin{figure}[t]
\centering
\includegraphics[width=\linewidth]{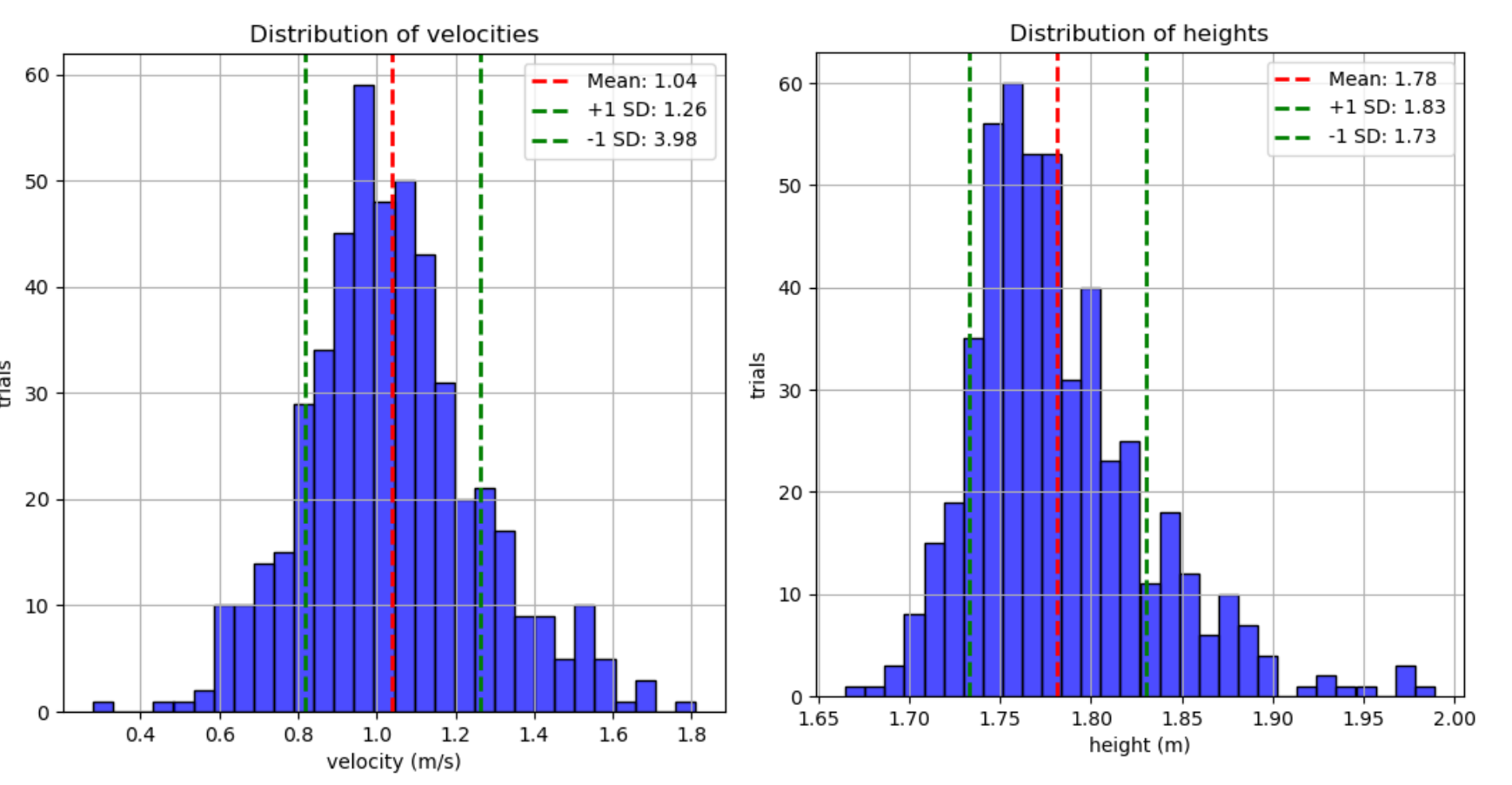}
\vspace*{-7mm}
\caption{Distribution of locomotion velocities and participants' heights in the DARKO dataset.}
\label{fig:darko_statistics}
\end{figure}

To further concentrate on human navigation activities typically encountered by robots in intralogistic environments, we collected the \textbf{DARKO} dataset using the truncation-robust, monocular MeTRAbs absolute 3D human pose estimator \cite{sarandi2020metrabs} with an Azure Kinect RGB \cite{azurekinect} input running at 16 Hz on our mobile robot, see Fig.~\ref{fig:cover}.
The DARKO dataset, collected from an egocentric view of a robot and therefore useful for human-aware navigation, makes a valuable addition to the limited set of datasets available from this perspective \cite{mangalam2020disentangling}. The DARKO dataset includes various locomotion modes, such as including slow walking, wavy walking, deviated navigation, and running, captured over 3 hours from 17 participants. We also highlight the diversity in participants' heights, as shown in Fig.~\ref{fig:darko_statistics}.
In summary, the DARKO dataset has 508 trials, each trial is $4.2\pm 1.15$ seconds. The mean trajectory length is $4.14 \pm 0.82$ meters, with velocity ranging from $0.6$ to $1.6$ m/s. For evaluation purposes, we select all trials from one actor and at least one trail from the other 16 actors for testing (5K frames), and use the rest for training (30K frames). The input horizon is set to 1.0 \si{\sec} (15 frames), and the output is 2.0 \si{\sec} (30 frames).

\textbf{Metrics:} We use Average and Final Displacement Errors (ADE/FDE), measured on the root joint (ADE\textsubscript{Tr}) and the pose joints (ADE\textsubscript{Po}) in meters. ADE is the mean square error between predicted and ground truth joint coordinates, calculated over the entire output sequence, while FDE measures displacement in the last time step only. When calculating ADE/FDE\textsubscript{Po}, to isolate pose prediction error, we subtract all other joints from the root joint to remove global translation. Furthermore, we report the runtime (R) in milliseconds to illustrate the duration of a single forward pass of the model with a batch size of 1. 


\subsection{Quantitative Results}
    
\begin{table}[t]
\renewcommand{\arraystretch}{1} 
\raggedleft
\begin{tabularx}{0.48\textwidth}{X p{1.5cm} p{1.5cm} p{0.7cm} p{0.7cm}} 
\toprule
\textbf{Method} & \textbf{ADE/FDE\textsubscript{Po} (m)$\downarrow$} & \textbf{ADE/FDE\textsubscript{Tr} (m)$\downarrow$} & \textbf{I (msec) $\downarrow$} & \textbf{R (msec) $\downarrow$} \\
\midrule
DLow \cite{yuan2020dlow} & 0.48 / 0.62 & 0.19 / 0.45 & 20 & 36.8\\
DMMGAN \cite{nikdel2023dmmgan} & \textbf{0.44} / \textbf{0.52} & \textbf{0.12} / \textbf{0.23} & 100 & 184\\
STPOTR \cite{mahdavian2023stpotr} & 0.50 / 0.75 & 0.13 / 0.27 & 25 & 46\\
UPTor & 0.51 / 0.74 & \textbf{0.12} / 0.25 & - & \textbf{17}\\
\bottomrule
\end{tabularx}
\caption[Baseline comparison on Human3.6M]{Evaluation on Human3.6M dataset across 15 distinct actions, including both navigational and static activities.}
\label{tab:human36m_table}
\vspace*{-5mm}
\end{table}
\begin{figure}[!t]
\centering
\includegraphics[width=\linewidth]{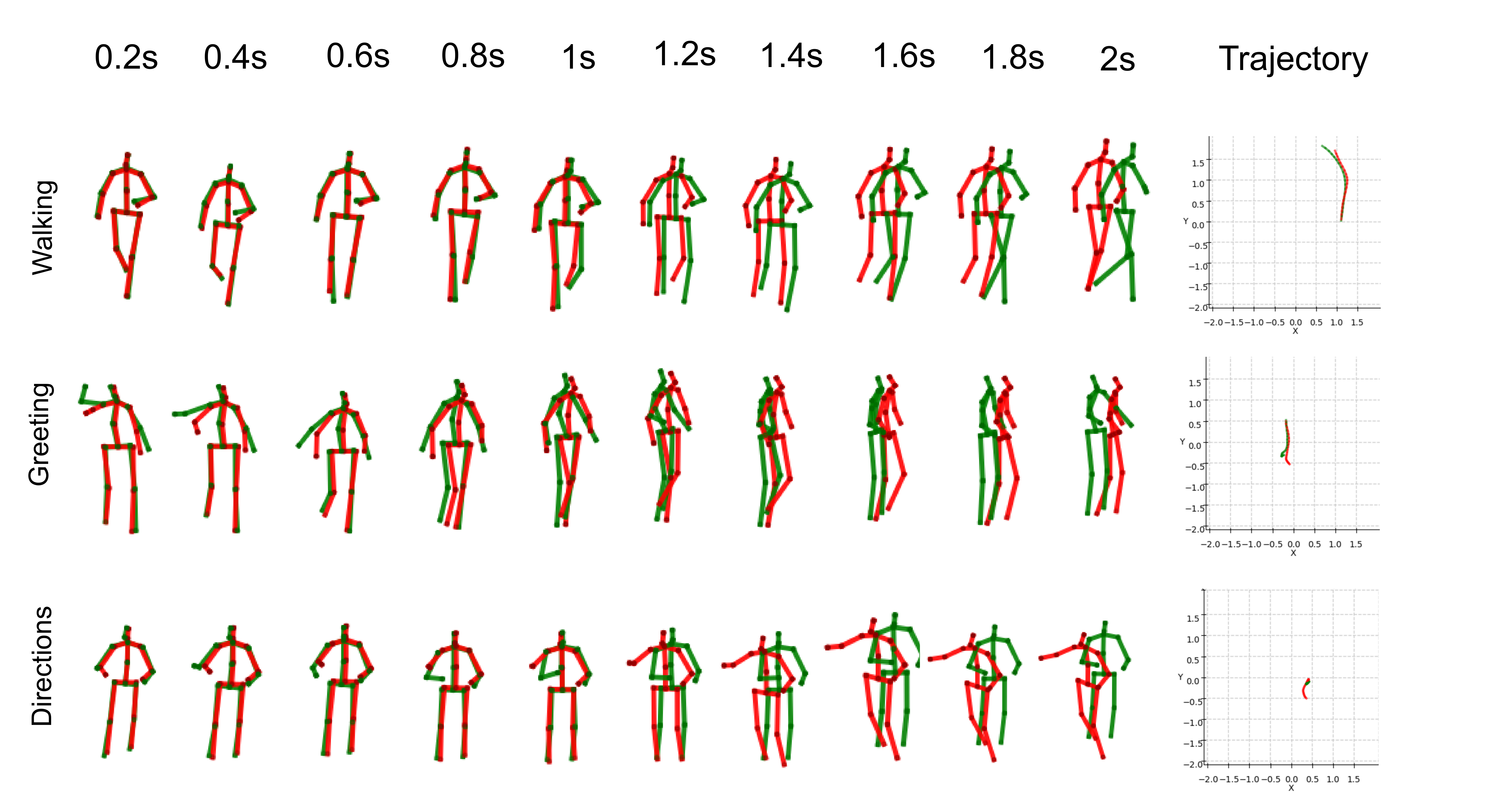}
\vspace*{-6mm}
\caption{Human 3.6M dataset predictions across 2-second horizon at 10 Hz}
\label{fig:h36m_dataset}
\vspace*{-2mm}
\end{figure}

\textbf{Human3.6M \cite{h36m_pami}:} Table \ref{tab:human36m_table} presents a quantitative comparison of our method, UPTor, against four established baselines \cite{mahdavian2023stpotr, yuan2020dlow, nikdel2023dmmgan}. The baseline values are taken from \cite{mahdavian2023stpotr}, and we make sure to strictly follow their evaluation setup. DLow excels at pose prediction but struggles with trajectory prediction, making it effective for static activities but less suitable for navigational activities that involve global translation. DMMGAN excels in both pose and trajectory prediction but its larger model size and longer runtime limit its applicability in real-time or practical robotic applications. It is important to note that the major part of the H3.6M evaluation set is comprised of static actions such as sitting, smoking, eating, or discussing. Given that these actions are heavily influenced by individual behavioral patterns, the superior performance of DLow and DMMGAN is due to their generative nature, which allows them to forecast a wide range of potential future poses/trajectories, and the error metric is computed between the closest generated sample and ground truth.

UPTor demonstrates comparable pose prediction and better trajectory prediction with a smaller network and better runtime compared to STPOTR across all 15 actions in the H3.6M dataset. STPOTR is a decoupled prediction method and uses two distinct module and for pose and prediction. Instead, we adopt a unified architecture and training process for both pose and trajectory prediction, thanks to our motion transformation module. The trajectories are predicted with high precision as they account for a significant portion of training error. In addition, our model uses pose dynamics as an added context for effective trajectory learning. To assess computational efficiency and model complexity, we compared the number of transformer parameters and conducted inference runtime tests on our model, UPTor, and STPOTR. While STPOTR contains 43,276,992 learnable transformer parameters, UPTor has 23,165,184 parameters. The runtime results in the I (msec) column are reported from the STPOTR paper based on their hardware. The R (msec) values for the UPTor and STPOTR models in Table \ref{tab:human36m_table} were obtained from our workstation, equipped with an AMD Ryzen Threadripper PRO 3995WX 64-Core CPU. Runtime values for other baselines in the R (msec) column are scaled accordingly with respect to the I (msec) column. In summary, the empirical results demonstrate that our model is capable of inferring human pose dynamics and trajectory patterns with a 50\% reduction in model size and improved runtime performance. 

\textbf{DARKO and CMU-Mocap \cite{cmumocap}:} 
To quantitatively compare performance of our model particularly on navigational sequences, we train both STPOTR and UPTor on our DARKO dataset and CMU-Mocap locomotion selected sequences. Quantitative results in Table \ref{tab:darko_cmu} indicate better performance of our model in predicting both poses and trajectories for navigating humans. 

\begin{table}[t]
\renewcommand{\arraystretch}{1} 
\centering
\begin{tabularx}{0.48\textwidth}{p{1.44cm} X X X X}
\toprule
\multirow{2}{*}{\textbf{Method}} & \multicolumn{2}{c}{DARKO} & \multicolumn{2}{c}{CMU-Mocap} \\
\cmidrule(lr){2-3} \cmidrule(lr){4-5}
 & \textbf{ADE/FDE\textsubscript{Po} (m)}$\downarrow$ & \textbf{ADE/FDE\textsubscript{Tr} (m)}$\downarrow$ & \textbf{ADE/FDE\textsubscript{Po} (m)}$\downarrow$ & \textbf{ADE/FDE\textsubscript{Tr} (m)}$\downarrow$ \\
\midrule
STPOTR \cite{mahdavian2023stpotr} & 0.47/0.65 & 0.17/0.34 & 0.58/0.86 & 0.09/0.18 \\
UPTor & \textbf{0.39}/\textbf{0.55} & \textbf{0.13}/\textbf{0.26} & \textbf{0.45}/\textbf{0.71} & \textbf{0.07}/\textbf{0.14} \\
\bottomrule
\end{tabularx}
\caption{Error metrics comparison on DARKO data \& locomotion actions subset of CMU-Mocap}
\label{tab:darko_cmu}
\vspace*{-6mm}
\end{table}

\begin{figure}[!t]
\centering
\includegraphics[width=\linewidth]{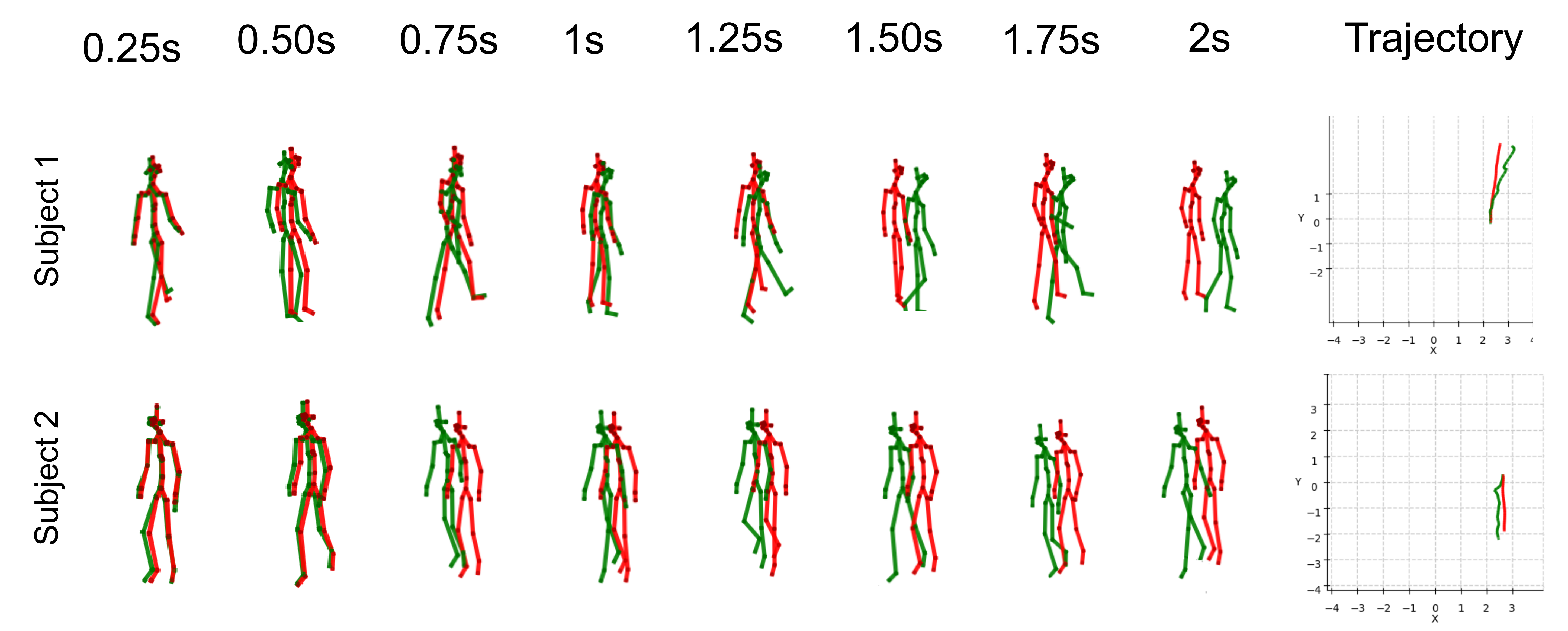}
\vspace*{-7mm}
\caption{DARKO dataset predictions across 2-second horizon at 16Hz}
\label{fig:darko_dataset}
\vspace*{-2mm}
\end{figure}

\begin{figure}[t]
\centering
\includegraphics[width=\linewidth]{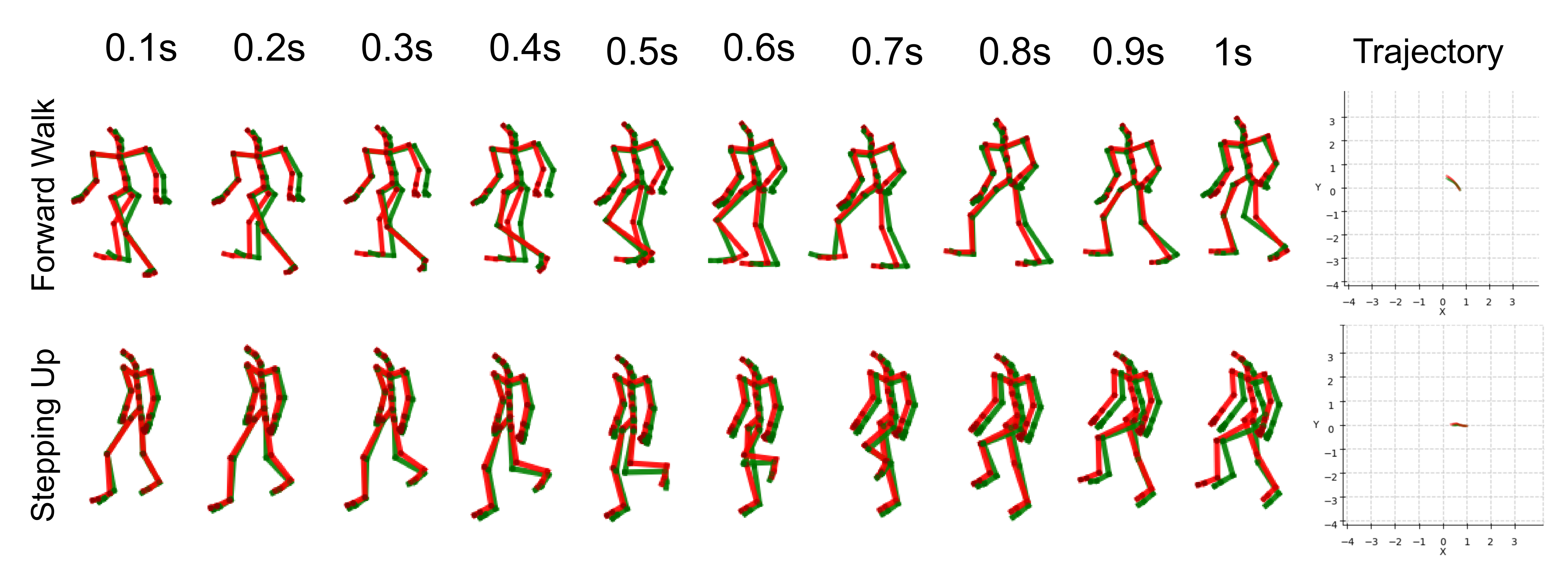}
\vspace*{-6mm}
\caption{CMU-Mocap dataset predictions across 1-second horizon at 10Hz}
\label{fig:cmu_dataset}
\vspace*{-1mm}
\end{figure}

\subsection{Qualitative Results}
Figs. ~\ref{fig:h36m_dataset}, ~\ref{fig:darko_dataset}, and ~\ref{fig:cmu_dataset} show frame-wise visualizations of predicted and ground-truth poses, along with top-down views of trajectory displacement in global coordinate space. The results are depicted from the start (left) to the end (right) of the prediction horizon for various actions and subjects along with their respective time stamps. Predicted poses are shown in red skeletons while the ground truth poses are in green. In H3.6M, our results are highly accurate up to the 1-second mark, with minor deviations thereafter. On DARKO, skeletons are depicted across two different subjects navigating in different directions. A closer observation of the trajectories reveals variance in motion patterns such as a wavy walk, a tendency to deviate slowly. The changes in motion speed are hinted at by trajectory lengths, given that we keep the prediction horizon constant. All of these motion patterns are captured well in UPTor predictions. The predicted poses also align well with the distinct ground truth skeletal features such as body inclination angle, hand swing, and so on. The predicted poses and trajectories from CMU-Mocap dataset show smooth motion transitions between frames as all the predicted frames are depicted. Although trajectories are short due to the short-term prediction, they exhibit diverse motion orientations. Thanks to our proposed transformation technique, predicted trajectories are well aligned with the ground truth sequences in spite of diverse motion orientations.

\subsection{Ablation Study}
\begin{table}[t]
\renewcommand{\arraystretch}{1} 
\centering
\begin{tabularx}{0.48\textwidth}{p{2.1cm} X X}
\toprule
\textbf{Test Cases} & \textbf{w/o transformation ADE\textsubscript{Po}/ADE\textsubscript{Tr} (m)$\downarrow$} & \textbf{with transformation ADE\textsubscript{Po}/ADE\textsubscript{Tr} (m)$\downarrow$} \\
\midrule
Original            & 0.46 / \textbf{0.13} & \textbf{0.39} / \textbf{0.13} \\
Translate           & 1.12 / 5.51 & \textbf{0.39} / \textbf{0.13} \\
Rotate              & 1.08 / 2.47 & \textbf{0.39} / \textbf{0.13} \\
Translate + Rotate  & 1.24 / 6.40 & \textbf{0.39} / \textbf{0.13} \\
\bottomrule
\end{tabularx}
\caption[Geometric Motion Robustness]{Evaluation on DARKO data under diverse spatial transformations and motion orientations.}
\label{tab:motionrobust_table}
\vspace*{-6mm}
\end{table}

\begin{table}[t]
\renewcommand{\arraystretch}{1}
\centering
\begin{tabularx}{0.48\textwidth}{p{2.7cm} X X }
\toprule
\textbf{Method} & \textbf{ADE/FDE\textsubscript{Po}(m)}$\downarrow$ & \textbf{ADE/FDE\textsubscript{Tr}(m)}$\downarrow$ \\
\midrule
without GAT & 0.42 / 0.59 & 0.13 / 0.27 \\
w/o masked self-attn & 0.40 / 0.57 & 0.13 / \textbf{0.25} \\
w/o shared attn & 0.43 / 0.58 & 0.13 / 0.27 \\
ours & \textbf{0.39} / \textbf{0.55} & \textbf{0.13} / 0.26 \\
\bottomrule
\end{tabularx}
\caption[Ablation study]{Comparison of error metrics across various model iterations on DARKO dataset}
\label{tab:ablation_table}
\vspace*{-6mm}
\end{table}


The trajectories from the DARKO dataset in Fig.~\ref{fig:darko_dataset} are centered around the same global space and motion orientation, as the robot was placed in a fixed position during data recording and subjects moved around a fixed global coordinate space. To demonstrate consistent performance of our model across various spatial coordinates and motion orientations, we trained the model using the original joints' spatial coordinates and motion orientation and tested its efficacy under four distinct scenarios outlined in Table \ref{tab:motionrobust_table}. In the ``Original'' scenario, motion sequences in the test set remain unchanged. For the ``Translate'' scenario, each sequence in test set undergoes random translation between \(-10\) and \(+10\) meters. In the ``Rotate'' scenario, motion direction is randomly rotated between \(-3.14\) and \(+3.14\) radians. In the ``Translate + Rotate'' scenario, sequences undergo both random rotation and translation. The model with motion transformation component show consistent results across all scenarios, substantiating the model's in-variance to global translation and rotation. Without motion transformation, results show good predictions on the original data as the model overfits to the recorded coordinate space and motion orientation, but struggles when it sees sequences that are subjected to random translations and rotations. Error values in Table \ref{tab:motionrobust_table} indicate that trajectories are more adversely impacted without transformation module. 

Table \ref{tab:ablation_table} examines the impact of other components in our model. In the first row, we exclude the graph attention network and employ a linear layer for embedding generation. In the second row, we utilize standard self-attention from \cite{vaswani2017attention}, replacing our model's specialized masked temporal self-attention with relative position representation. In the third row, we remove the shared attention component. Finally, in the last row, we show our proposed model. Error values indicate that pose prediction accuracy is improved by including these model components. 

\subsection{Limitations}
In this paper we propose to predict trajectories of navigating humans jointly with full-body poses in global coordinate space. Due to our specific focus on navigation activities, the method does not reach state of the art performance in static activities and fine-grained joints movements, such as gestures or interactions with objects, compared to the state of the art methods dedicated to such activities. 
Furthermore, we noticed practical challenges in deploying existing prediction method on a robot under increased workload, which leads to missing frames from camera sensor and results in variable input frequencies to the transformer.

\section{Conclusions}
\label{sec:conclusions}
In summary, our method reaches an appealing accuracy, model size, and runtime balance for pose and trajectory prediction in scenarios which are most critical for human-aware navigation of a mobile robot. 
For future work, we plan to address real-world deployment challenges such as missing frames and varying perception frequencies. We also aim to incorporate contextual cues from static and dynamic environments to enhance prediction accuracy \cite{rudenko2020human}. Finally, we intend to extend the model to a multi-person setting.










\clearpage
\printbibliography
\end{document}